\documentclass[11pt]{article}
\usepackage{coling2016}
\usepackage{times}
\usepackage{url}
\usepackage{latexsym}

\usepackage{latexsym}
\usepackage{graphicx}
\usepackage{amssymb}
\usepackage{subfigure}
\usepackage{multirow}
\usepackage{graphicx}
\usepackage{epstopdf}
\usepackage{amsmath}
\usepackage{url}

\title{Effective LSTMs for Target-Dependent Sentiment Classification}

\author{Duyu Tang, Bing Qin, Xiaocheng Feng,  Ting Liu\\
	Harbin Institute of Technology, Harbin, China\\
	\{dytang,\ qinb,\ xcfeng, \ tliu\}@ir.hit.edu.cn}

\date{}

\begin{document}
\maketitle
\begin{abstract}
  Target-dependent sentiment classification remains a challenge: modeling the semantic relatedness of a target with its context words in a sentence.
  Different context words have different influences on determining the sentiment polarity of a sentence towards the target.
  Therefore, it is desirable to integrate the connections between target word and context words when building a learning system.
  In this paper, we develop two target dependent long short-term memory (LSTM) models, where target information is automatically taken into account.
  We evaluate our methods on a benchmark dataset from Twitter.
  Empirical results show that modeling sentence representation with standard LSTM does not perform well.
  Incorporating target information into LSTM can significantly boost the classification accuracy.
  The target-dependent LSTM models achieve state-of-the-art performances without using syntactic parser or external sentiment lexicons.\footnote{Codes are publicly available at \url{http://ir.hit.edu.cn/~dytang}.}
\end{abstract}

\section{Introduction}

\blfootnote{
	%
	% for review submission
	%
%	\hspace{-0.65cm}  % space normally used by the marker
%	Place licence statement here for the camera-ready version, see
%	Section~\ref{licence} of the instructions for preparing a
%	manuscript.
	
	 % final paper: en-uk version 
	
%	 \hspace{-0.65cm}  % space normally used by the marker
%	 This work is licenced under a Creative Commons 
%	 Attribution 4.0 International Licence.
%	 Licence details:
%	 \url{http://creativecommons.org/licenses/by/4.0/}
	 
	 % final paper: en-us version 
	
	 \hspace{-0.65cm}  % space normally used by the marker
	 This work is licensed under a Creative Commons 
	 Attribution 4.0 International License.
	 License details:
	 \url{http://creativecommons.org/licenses/by/4.0/}
}

Sentiment analysis, also known as opinion mining \cite{Pang2008,Liu2012a}, is a fundamental task in natural language processing and computational linguistics. Sentiment analysis is crucial to understanding user generated text in social networks or product reviews, and has drawn a lot of attentions from both industry and academic communities. 
In this paper, we focus on target-dependent sentiment classification \cite{Jiang2011,Dong2014a,Vo2015}, which is a fundamental and extensively studied task in the field of sentiment analysis. 
Given a sentence and a target mention, the task calls for inferring the sentiment polarity (e.g. positive, negative, neutral) of the sentence towards the target. For example, let us consider the sentence: ``\textit{I bought a new camera. The {picture quality} is amazing but the {battery life} is too short}''. If the target string is \textit{\underline{picture quality}}, the expected sentiment polarity is ``positive'' as the sentence expresses a positive opinion towards \textit{picture quality}. If we consider the target as \textit{\underline{battery life}}, the correct sentiment polarity should be ``negative''.

Target-dependent sentiment classification is typically regarded as a kind of text classification problem in literature.
Majority of existing studies build sentiment classifiers with supervised machine learning approach, such as feature based Supported Vector Machine \cite{Jiang2011} or neural network approaches \cite{Dong2014a,Vo2015}.
Despite the effectiveness of these approaches, we argue that target-dependent sentiment classification remains a challenge: how to effectively model the semantic relatedness of a target word with its context words in a sentence. 
One straight forward way to address this problem is to manually design a set of target-dependent features, and integrate them into existing feature-based SVM. 
However, feature engineering is labor intensive and the ``sparse'' and ``discrete'' features are clumsy in encoding side information like target-context relatedness.
In addition, a person asked to do this task will naturally ``look at'' parts of relevant context words which are helpful to determine the sentiment polarity of a sentence towards the target. 
These motivate us to develop a powerful neural network approach, which is capable of learning continuous features (representations) without feature engineering and meanwhile capturing the intricate relatedness between target and context words.

In this paper, we present neural network models to deal with target-dependent sentiment classification. 
The approach is an extension on long short-term memory (LSTM) \cite{Hochreiter1997} by incorporating target information. 
Such target-dependent LSTM approach models the relatedness of a target word with its context words, and selects the relevant parts of contexts to infer the sentiment polarity towards the target. 
The model could be trained in an end-to-end way with standard backpropagation, where the loss function is cross-entropy error of supervised sentiment classification.

We apply the neural model to target-dependent sentiment classification on a benchmark dataset \cite{Dong2014a}. We compare with feature-based SVM \cite{Jiang2011}, adaptive recursive neural network \cite{Dong2014a} and lexicon-enhanced neural network \cite{Vo2015}.
Empirical results show that the proposed approach without using syntactic parser or external sentiment lexicon obtains state-of-the-art classification accuracy.
In addition, we find that modeling sentence with standard LSTM does not perform well on this target-dependent task.
Integrating target information into LSTM could significantly improve the classification accuracy.
%The main contributions of this work are as follows:
%
%\begin{itemize}
%	\item We develop target-dependent LSTM models for target-dependent sentiment classification, where the relatedness of target word with context words is encoded.
%	\item Empirical results show that the neural model obtains state-of-the-art performance on a benchmark dataset. Incorporating target information into LSTM could improve the classification accuracy.
%\end{itemize}

%\section{The Approach}
\section{The Approach}
We describe the proposed approach for target-dependent sentiment classification in this section. 
We first present a basic long short-term memory (LSTM) approach, which models the semantic representation of a sentence without considering the target word being evaluated.
Afterwards, we extend LSTM by considering the target word, obtaining the Target-Dependent Long Short-Term Memory (TD-LSTM) model.
Finally, we extend TD-LSTM with target connection, where the semantic relatedness of target with its context words are incorporated.

\begin{figure*}[t]
	\centering
	\includegraphics[width=.95\textwidth]{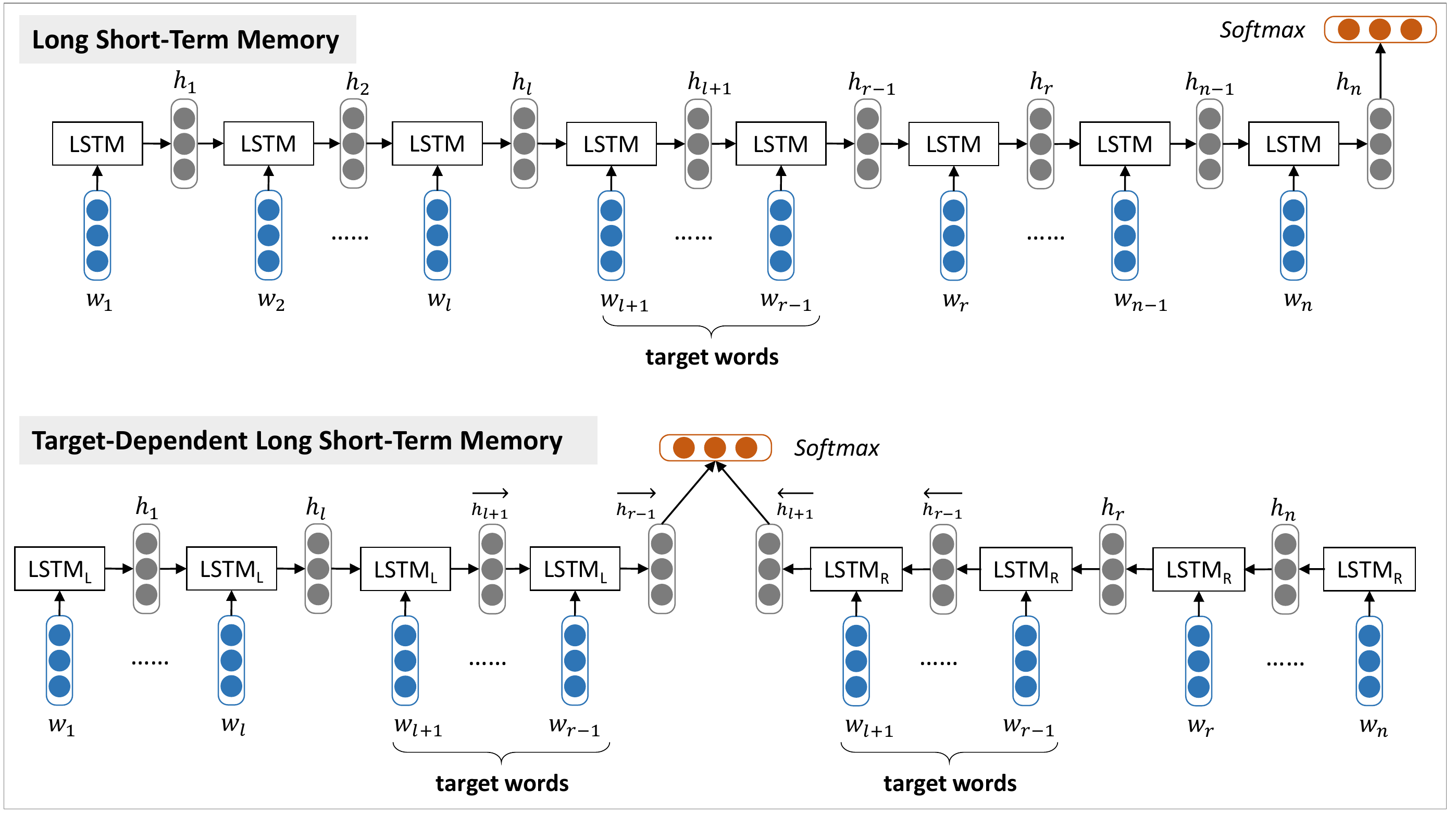}
	\caption{The basic long short-term memory (LSTM) approach and its target-dependent extension TD-LSTM for target-dependent sentiment classification. $w$ stands for word in a sentence whose length is $n$, \{$w_{l+1}$, $w_{l+2}$, ..., $w_{r-1}$\} are target words, \{$w_{1}$, $w_{2}$, ..., $w_{l}$\} are preceding context words, \{$w_{r}$, ..., $w_{n-1}$, $w_n$\} are following context words.}
	\label{fig:lstm-bilstm}
\end{figure*}

\begin{figure*}[t]
	\centering
	\includegraphics[width=.95\textwidth]{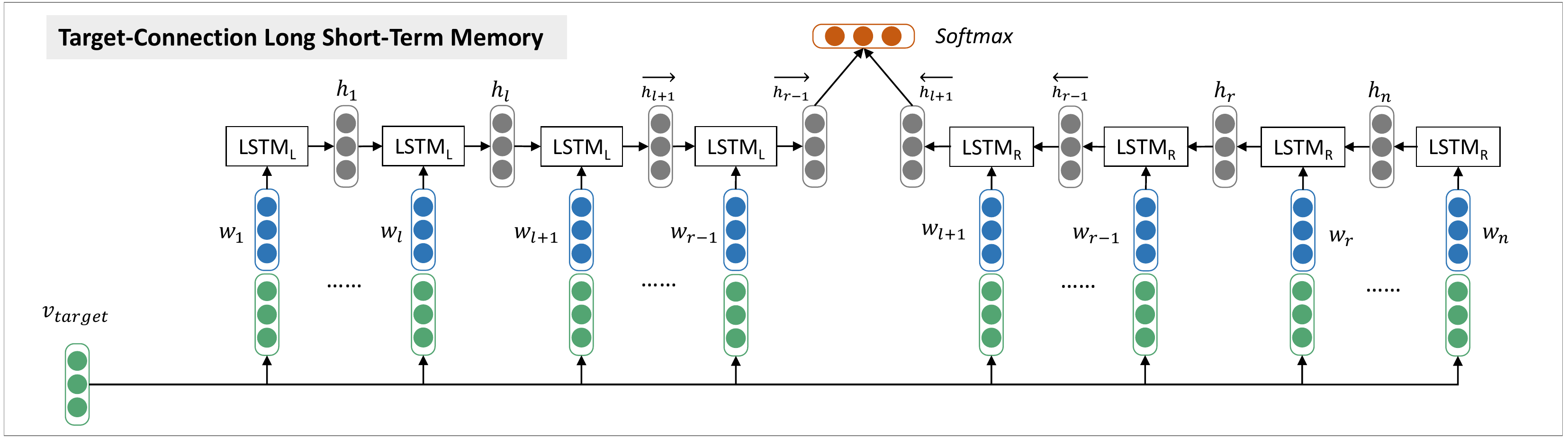}
	\caption{The target-connection long short-term memory (TC-LSTM) model for target-dependent sentiment classification, where $w$ stands for word in a sentence whose length is $n$, \{$w_{l+1}$, $w_{l+2}$, ..., $w_{r-1}$\} are target words, $v_{target}$ is target representation, \{$w_{1}$, $w_{2}$, ..., $w_{l}$\} are preceding context words, \{$w_{r}$, ..., $w_{n-1}$, $w_n$\} are following context words.}
	\label{fig:framework}
\end{figure*}

\subsection{Long Short-Term Memory (LSTM)}
In this part, we describe a long short-term memory (\textbf{LSTM}) model for target-dependent sentiment classification.
It is a basic version of our approach.
In this setting, the target to be evaluated is ignored so that the task is considered in a target independent way. 

We use LSTM as it is a state-of-the-art performer for semantic composition in the area of sentiment analysis \cite{Li2015a,Tang2015}. 
It is capable of computing the representation of a longer expression (e.g. a sentence) from the representation of its children with multi levels of abstraction.
The sentence representation can be naturally considered as the feature to predict the sentiment polarity of sentence. 

Specifically, each word is represented as a low dimensional, continuous and real-valued vector, also known as word embedding \cite{Bengio2003,Mikolov2013a,Pennington2014,Tang2014}.
All the word vectors are stacked in a word embedding matrix $L_w \in \mathbb{R}^{d \times |V|}$, where $d$ is the dimension of word vector and $|V|$ is vocabulary size.
In this work, we pre-train the values of word vectors from text corpus with embedding learning algorithms \cite{Pennington2014,Tang2014} to make better use of semantic and grammatical associations of words. 
%The comparison between different embedding learning algorithms is given in the experiment section. 

We use LSTM to compute the vector of a sentence from the vectors of words it contains, an illustration of the model is shown in Figure \ref{fig:lstm-bilstm}. LSTM is a kind of recurrent neural network (RNN), which is capable of mapping vectors of words with variable length to a fixed-length vector by recursively transforming current word vector $w_t$ with the output vector of the previous step $h_{t-1}$. 
The transition function of standard RNN is a linear layer followed by a pointwise non-linear layer such as hyperbolic tangent function~($tanh$).
\begin{equation}
h_t = tanh(W \cdot [h_{t-1};w_{t}] + b)
\end{equation}
where $W \in \mathbb{R}^{d \times 2d }$, $b \in \mathbb{R}^{d}$, $d$ is dimension of word vector. 
However, standard RNN suffers the problem of gradient vanishing or exploding \cite{Bengio1994,Hochreiter1997}, where gradients may grow or decay exponentially over long sequences.
Many researchers use a more sophisticated and powerful LSTM cell as the transition function, so that long-distance semantic correlations in a sequence could be better modeled. 
Compared with standard RNN, LSTM cell contains three additional neural gates: an input gate, a forget gate and an output gate. 
These gates adaptively remember input vector, forget previous history and generate output vector \cite{Hochreiter1997}. LSTM cell is calculated as follows.
\begin{align} 
& i_t = \sigma(W_i \cdot [h_{t-1};w_{t}] + b_i)\\
& f_t = \sigma(W_f \cdot [h_{t-1};w_{t}] + b_f)\\
& o_t = \sigma(W_o \cdot [h_{t-1};w_{t}] + b_o)\\
& g_t = tanh(W_r \cdot [h_{t-1};w_{t}] + b_r)\\
& c_t = i_t \odot g_t + f_t \odot c_{t-1}\\
& h_t = o_t \odot tanh(c_t)
\end{align}
where $\odot$ stands for element-wise multiplication, $\sigma$ is sigmoid function, $W_i$, $b_i$, $W_f$, $b_f$, $W_o$, $b_o$ are the parameters of input, forget and output gates.

After calculating the hidden vector of each position, we regard the last hidden vector as the sentence representation \cite{Li2015a,Tang2015}.
We feed it to a linear layer whose output length is class number, and add a $softmax$ layer to output the probability of classifying the sentence as positive, negative or neutral. 
Softmax function is calculated as follows, where $C$ is the number of sentiment categories. 
\begin{equation}\label{equa:softmax}
\centering
softmax_i =\frac{exp(x_i)}{\sum_{i'=1}^Cexp(x_{i'})}
\end{equation}

\subsection{Target-Dependent LSTM (TD-LSTM)}
The aforementioned LSTM model solves target-dependent sentiment classification in a target-independent way. 
That is to say, the feature representation used for sentiment classification remains the same without considering the target words. 
Let us again take ``\textit{I bought a new camera. The {picture quality} is amazing but the {battery life} is too short}'' as an example.
The representations of this sentence with regard to \textit{\underline{picture quality}} and \textit{\underline{battery life}} are identical. This is evidently problematic as the sentiment polarity labels towards these two targets are different.

To take into account of the target information, we make a slight modification on the aforementioned LSTM model and introduce a target-dependent LSTM (\textbf{TD-LSTM}) in this subsection.
The basic idea is to model the preceding and following contexts surrounding the target string, so that contexts in both directions could be used as feature representations for sentiment classification. We believe that capturing such target-dependent context information could improve the accuracy of target-dependent sentiment classification. 

Specifically, we use two LSTM neural networks, a left one LSTM$_L$ and a right one LSTM$_R$, to model the preceding and following contexts respectively. An illustration of the model is shown in Figure \ref{fig:lstm-bilstm}.
The input of LSTM$_L$ is the preceding contexts plus target string, and the input of LSTM$_R$ is the following contexts plus target 
string.
We run LSTM$_L$ from left to right, and run LSTM$_R$ from right to left.
We favor this strategy as we believe that regarding target string as the last unit could better utilize the semantics of target string when using the composed representation for sentiment classification.
Afterwards, we concatenate the last hidden vectors of LSTM$_L$ and LSTM$_R$, and feed them to a $softmax$ layer to classify the sentiment polarity label.
One could also try averaging or summing the last hidden vectors of LSTM$_L$ and LSTM$_R$ as alternatives.

\subsection{Target-Connection LSTM (TC-LSTM)}
Compared with LSTM model, target-dependent LSTM (TD-LSTM) could make better use of the target information.
However, we think TD-LSTM is still not good enough because it does not capture the interactions between target word and its contexts.
Furthermore, a person asked to do target-dependent sentiment classification will select the relevant context words which are helpful to determine the sentiment polarity of a sentence towards the target. 

Based on the consideration mentioned above, we go one step further and develop a target-connection long short-term memory (\textbf{TC-LSTM}). 
This model extends TD-LSTM by incorporating an target connection component, which explicitly utilizes the connections between target word and each context word when composing the representation of a sentence.

An overview of TC-LSTM is illustrated in Figure \ref{fig:framework}. 
The input of TC-LSTM is a sentence consisting of $n$ words $\{w_1, w_2, ... w_n\}$ and a target string $t$ occurs in the sentence. We represent target $t$ as $\{w_{l+1}, w_{l+2} ...w_{r-1}\}$ because a target could be a word sequence of variable length, such as ``\textit{google}'' or ``\textit{harry potter}''. 
When processing a sentence, we split it into three components: target words, preceding context words and following context words.
We obtain target vector $v_{target}$ by averaging the vectors of words it contains, which has been proven to be simple and effective in representing named entities \cite{Socher2013b,Sun2015}. 
When compute the hidden vectors of preceding and following context words, we use two separate long short-term memory models, which are similar with the strategy used in TD-LSTM.
The difference is that in TC-LSTM the input at each position is the concatenation of word embedding and target vector $v_{target}$, while in TD-LSTM the input at each position only includes the embedding of current word. 
We believe that TC-LSTM could make better use of the connection between target and each context word when building the representation of a sentence. 

\subsection{Model Training}
We train LSTM, TD-LSTM and TC-LSTM in an end-to-end way in a supervised learning framework. 
The loss function is the cross-entropy error of sentiment classification.

\begin{equation}
loss = -\sum_{s \in S}^{}\sum_{c = 1}^{C}P_{c}^{g}(s) \cdot log(P_{c}(s))
\end{equation}
where $S$ is the training data, $C$ is the number of sentiment categories, $s$ means a sentence, $P_c(s)$ is the probability of predicting $s$ as class $c$ given by the $softmax$ layer, $P^g_c(s)$ indicates whether class $c$ is the correct sentiment category, whose value is 1 or 0.
We take the derivative of loss function through back-propagation with respect to all parameters, and update parameters with stochastic gradient descent.

\section{Experiment}
We apply the proposed method to target-dependent sentiment classification to evaluate its effectiveness.
We describe experimental setting and empirical results in this section.

\subsection{Experimental Settings}
We conduct experiment in a supervised setting on a benchmark dataset \cite{Dong2014a}. Each instance in the training/test set has a manually labeled sentiment polarity. 
Training set contains 6,248 sentences and test set has 692 sentences. 
The percentages of positive, negative and neutral in training and test sets are both 25\%, 25\%, 50\%.
We train the model on training set, and evaluate the performance on test set. 
Evaluation metrics are accuracy and macro-F1 score over positive, negative and neutral categories \cite{Manning1999,Jurafsky2000}.

\subsection{Comparison to Other Methods}

We compare with several baseline methods, including:

In \textbf{SVM-indep}, SVM classifier is built with target-independent features, such as unigram, bigram, punctuations, emoticons, hashtags, the numbers of positive or negative words in General Inquirer sentiment lexicon. 
In \textbf{SVM-dep}, target-dependent features \cite{Jiang2011} are also concatenated as the feature representation. 

In \textbf{Recursive NN}, standard Recursive neural network is used for feature learning over a transfered target-dependent dependency tree \cite{Dong2014a}. \textbf{AdaRNN-w/oE}, \textbf{AdaRNN-w/E} and \textbf{AdaRNN-comb} are different variations of adaptive recursive neural network \cite{Dong2014a}, whose composition functions are adaptively selected according to the inputs. 

In \textbf{Target-dep}, SVM classifier is built based on rich target-independent and target-dependent features \cite{Vo2015}. In \textbf{Target-dep$^+$}, sentiment lexicon features are further incorporated. 

The neural models developed in this paper are abbreviated as LSTM, TD-LSTM and TC-LSTM, which are described in the previous section. We use 100-dimensional Glove vectors learned from Twitter, randomize the parameters with uniform distribution $U(-0.003, 0.003)$, set the clipping threshold of softmax layer as 200 and set learning rate as 0.01.

\begin{table}[h]
	\centering
	\begin{tabular}{l|cc}
		\hline
		Method & Accuracy & Macro-F1\\
		\hline
		SVM-indep 			& 0.627 & 0.602 \\
		SVM-dep	 			& 0.634 & 0.633 \\
		Recursive NN		& 0.630 & 0.628 \\
		AdaRNN-w/oE		 	& 0.649 & 0.644 \\
		AdaRNN-w/E	 		& 0.658 & 0.655 \\
		AdaRNN-comb			& 0.663 & 0.659 \\
		Target-dep			& 0.697 & 0.680 \\
		Target-dep$^+$ 		& 0.711 & \textbf{0.699} \\
		\hline
		LSTM 				& 0.665	& 0.647	\\
		TD-LSTM				& 0.708	& 0.690 \\
		TC-LSTM				& \textbf{0.715} & 0.695\\
		\hline
	\end{tabular}
	\caption{Comparison of different methods on target-dependent sentiment classification. Evaluation metrics are accuracy and macro-F1. Best scores are in {bold}.}
	\label{table:experiment-baseline}
\end{table}

Experimental results of baseline models and our methods are given in Table \ref{table:experiment-baseline}. 
Comparing between SVM-indep and SVM-dep, we can find that incorporating target information can improve the classification accuracy of a basic SVM classifier. 
AdaRNN performs better than feature based SVM by making use of dependency parsing information and tree-structured semantic composition.
We can find that target-dep is a strong performer even without using lexicon features. It benefits from rich automatic features generated from word embeddings. 
%By incorporating sentiment lexicon features, target-dep$^+$ obtains better performance.

Among LSTM based models described in this paper, the basic LSTM approach performs worst. This is not surprising because this task requires understanding target-dependent text semantics,  while the basic LSTM model does not capture any target information so that it predicts the same result for different targets in a sentence. 
TD-LSTM obtains a big improvement over LSTM when target signals are taken into consideration. This result demonstrates the importance of target information for target-dependent sentiment classification. 
By incorporating target-connection mechanism, TC-LSTM obtains the best performances and outperforms all baseline methods in term of classification accuracy. 

Comparing between Target-dep$^+$ and Target-dep, we find that sentiment lexicon feature could further improve the classification accuracy. Our final model TC-LSTM without using sentiment lexicon information performs comparably with Target-dep$^+$. We believe that incorporation lexicon information in TC-LSTM could get further improvement. We leave this as a potential future work. 

\subsection{Effects of Word Embeddings}
It is well accepted that a good word embedding is crucial to composing a powerful text representation at higher level. 
We therefore study the effects of different word embeddings on LSTM, TD-LSTM and TC-LSTM in this part. 
Since the benchmark dataset from \cite{Dong2014a} comes from Twitter, we compare between sentiment-specific word embedding (SSWE)\footnote{SSWE vectors are publicly available at \url{http://ir.hit.edu.cn/~dytang}} \cite{Tang2014} and Glove vectors\footnote{Glove vectors are publicly available at \url{http://nlp.stanford.edu/projects/glove/}} \cite{Pennington2014}. All these word vectors are 50-dimensional and learned from Twitter. 
SSWE$_h$, SSWE$_r$ and SSWE$_u$ are different embedding learning algorithms introduced in \cite{Tang2014}. SSWE$_h$ and SSWE$_r$ learn word embeddings by only using sentiment of sentences. SSWE$_u$ takes into account of sentiment of sentences and contexts of words simultaneously.

\begin{figure}[h]
	\centering
	\includegraphics[width=.62\textwidth]{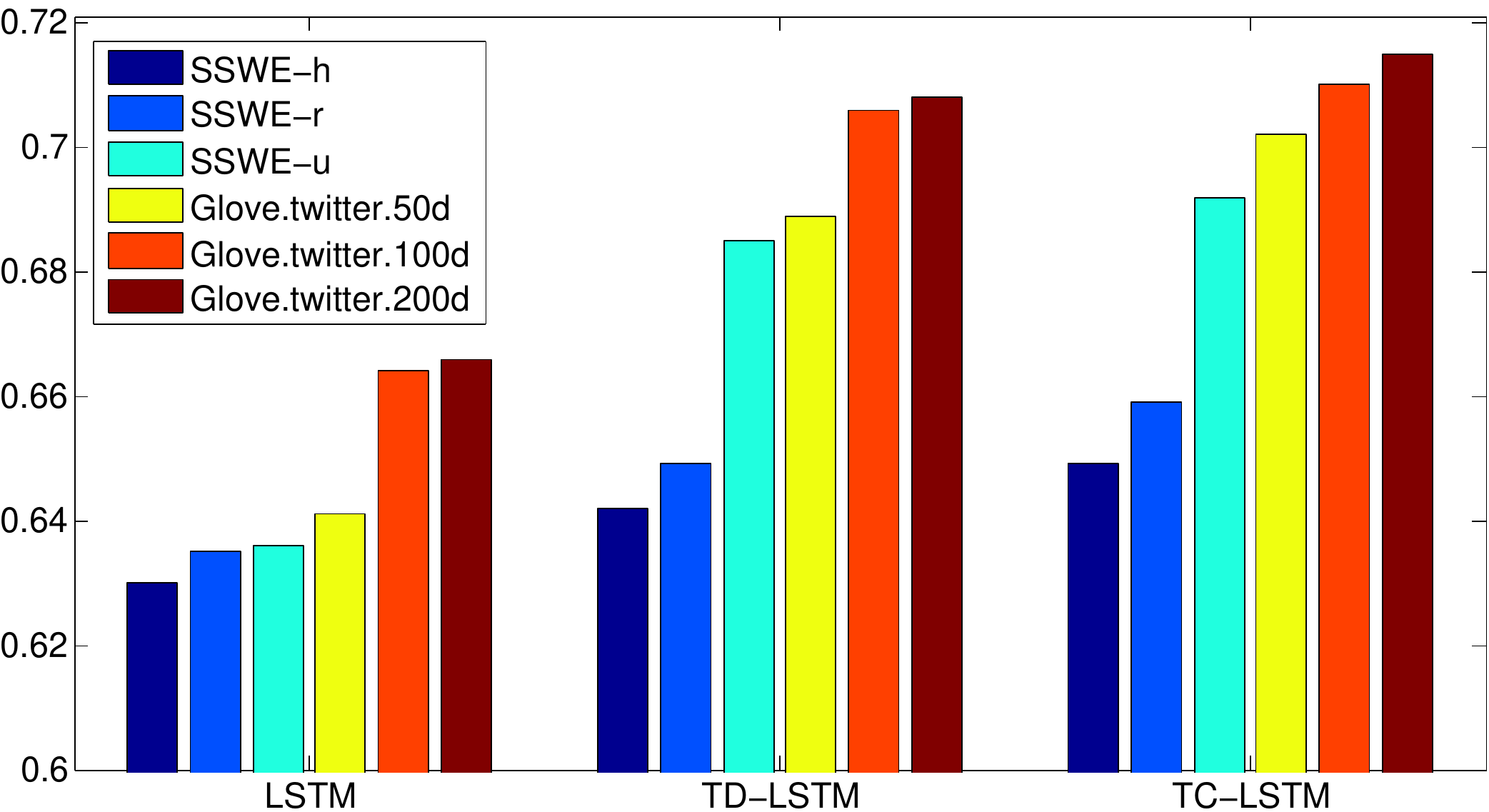}
	\caption{Classification accuracy of LSTM, TD-LSTM and TC-LSTM with different word embeddings. We compare between SSWE$_h$, SSWE$_r$, SSWE$_u$ and Glove vectors.}
	\label{fig:result-embedding}
\end{figure}
From Figure \ref{fig:result-embedding}, we can find that SSWE$_h$ and SSWE$_r$ perform worse than SSWE$_u$, which is consistent with the results reported on target-independent sentiment classification of tweets \cite{Tang2014}. This shows the importance of context information for word embedding learning as both  SSWE$_h$ and SSWE$_r$ do not encode any word contexts. 
Glove and SSWE$_u$ perform comparably, which indicates the importance of global context for estimating a good word representation. 
In addition, the target connection model TC-LSTM performs best when considering a specific word embedding.

%\begin{figure}[h]
%	\centering
%	\includegraphics[width=.45\textwidth]{pic/dms.pdf}
%	\caption{Classification accuracy of LSTM, TD-LSTM and TC-LSTM with Glove vectors of 50/100/200 dimensions.}
%	\label{fig:result-embedding-dms}
%\end{figure}

\begin{table}[h]
	\centering
	\begin{tabular}{l|c|c|c}
		\hline
		& 50dms & 100dms & 200dms\\
		\hline
		LSTM	& 	27	&	95		&  329\\
		LSTM-TD	&	20	&	93		&  274\\
		LSTM-TC	&	65	&	280		&  1,165\\
		\hline		
	\end{tabular}
	\caption{Time cost of each model with 50dms, 100dms and 200dms Glove vectors. Each value means how many seconds cost in each training iteration.}
	\label{table:embedding-dms}
\end{table}
We  compare between Glove vectors with different dimensions (50/100/200). Classification accuracy and time cost are given in Figure \ref{fig:result-embedding} and Table~\ref{table:embedding-dms}, respectively.
We can find that 100-dimensional word vectors perform better than 50-dimensional word vectors, while 200-dimensional word vectors do not show significant improvements. 
Furthermore, TD-LSTM and LSTM have similar time cost, while TD-LSTM gets higher classification accuracy as target information is incorporated. 
TC-LSTM performs slightly better than TD-LSTM while at the cost of longer training time because the parameter number of TC-LSTM is larger.

\subsection{Case Study}
In this section, we explore to what extent the target-dependent LSTM models including TD-LSTM and TC-LSTM improve the performance of a basic LSTM model.

%the difference between LSTM and target related LSTMs 
%\begin{table}\small
%	\centering
%	\subtable[LSTM ]{
%		\centering
%	\begin{tabular}{l|c|c|c}
%		\hline
%		& p=-1 & p=0 & p=1\\
%		\hline
%		g=-1	& 	110	&	58		&  5\\
%		g=0	&	41	&	258		&  47\\
%		g=1	&	26	&	52		&  95\\
%		\hline		
%	\end{tabular}
%	\label{table:error-table-lstm} }
%	\subtable[TD-LSTM ]{
%	\centering
%	\begin{tabular}{l|c|c|c}
%		\hline
%		& p=-1 & p=0 & p=1\\
%		\hline
%		g=-1& 	127	&	43		&  3\\
%		g=0	&	43	&	52		&  30\\
%		g=1	&	28	&	52		&  93\\
%		\hline		
%	\end{tabular}
%	\label{table:error-table-td-lstm} }
%\caption{}
%\label{key}
%\end{table}

\begin{table}[h]
	\centering
	\begin{tabular}{p{12cm}|c|c}
		\hline
		Example & gold  &  LSTM \\
		\hline
		\textit{i hate my \textbf{ipod} look at my last tweet before the argh one that 's for you}	& 	-1	&	0	 \\
		\hline
		\textit{okay soooo ... ummmmm .... what is going on with \textbf{lindsay lohan}' s face? boring day at the office = perez and tomorrow overload. not good} & 0 & -1 \\
		\hline
		\textit{i heard ShannonBrown did his thing in the \textbf{lakers} game!! got ta love him} & 0 & 1 \\
		\hline
		%		\textit{I hate when people think its weird that I listen to \textbf{taylor swift} , paramore , or anything pop \& rock ... am I not a pop \& rock type a chick?} & 1 & -1\\
		%		\hline
		%		\textit{oh , shucks! i 've got a stupid air bubble behind my \textbf{ipod} screen.} & 0 & -1\\
		%		\hline
		\textit{Hey \textbf{google}, thanks for all these great Labs features on Chromium, but how about '' Create Application Shortcut''?!} & 1 & 0 \\
		%		\hline
		%		\textit{Is it just me, or does \textbf{john boehner} sound like a newsman? Sounds like he belongs on CBS Nightly News.} & -1 & 0\\
		\hline
	\end{tabular}
	\caption{Examples drawn from the test set whose polarity labels are incorrectly inferred by LSTM but correctly predicted by both TD-LSTM and TC-LSTM. For each example, target words are in \textbf{bold}, ``gold'' is the ground truth and ``LSTM'' means the predicted sentiment label from LSTM model.}
	\label{table:case}
\end{table}

In Table \ref{table:case}, we list some examples whose polarity labels are incorrectly inferred by LSTM but correctly predicted by both TD-LSTM and TC-LSTM.
We observe that LSTM model prefers to assigning the polarity of the entire sentence while ignoring the target to be evaluated.
TD-LSTM and TC-LSTM could take into account of target information to some extend. 
For example, in the 2nd example the opinion holder expresses a negative opinion about his work, but holds a neutral sentiment towards the target ``\textit{lindsay lohan}''.
%In the 4th sentence the opinion holder expresses a negative opinion about how others think about him. However, he holds a positive opinion towards the target ``\textit{taylor swift}''.
In the last example, the whole sentence expresses a neutral sentiment while it holds a positive opinion towards ``\textit{google}''.

We analyse the error cases that both TD-LSTM and TC-LSTM cannot well handle, and find that 85.4\% of the misclassified examples relate to neutral category. The positive instances are rarely misclassified as negative, and vice versa. 
A example of errors is: ``\textit{freaky friday on television reminding me to think wtf happened to \textbf{lindsay lohan}, she was such a terrific actress , + my huge crush on haley hudson.}'', which is incorrectly predicted as positive towards target ``\textit{indsay lohan}'' in both TD-LSTM and TC-LSTM.

\subsection{Discussion}
In order to capture the semantic relatedness between target and context words, we extend TD-LSTM by adding a target connection component. One could also try other extensions to capture the connection between target and context words. For example, we also tried an attention-based LSTM model, which is inspired by the recent success of attention-based neural network in machine
translation \cite{Bahdanau2015} and document encoding \cite{Li2015}. 
%The basic idea of attention neural model is to choose ``where to look'' by assigning a weight/importance/probability to each input position when computing a upper level representation.
%Therefore, we consider target word as the ``key'' to assign the weight of its surrounding context word. We think this is a natural choice as target word is the one to be evaluated in a sentence.
We implement the soft-attention mechanism \cite{Bahdanau2015} to enhance TD-LSTM.
%\begin{figure}[h]
%	\centering
%	\includegraphics[width=.45\textwidth]{pic/attention.pdf}
%	\caption{An attention mechanism to assign an importance score for each context word. $v_{target}$ is the vector representation, $h_i$ stands for the hidden vector of $i$-th context word.}
%	\label{fig:attention}
%\end{figure}
We incorporate two attention layers for preceding LSTM and following LSTM, respectively.
The output vector for each attention layer is the weighted average among hidden vectors of LSTM, 
%namely $vec = \sum_{i=1}^{k}\alpha_i h_i$.
where the weight of each hidden vector is calculated with a feedforward neural network.
The outputs of preceding and following attention models are concatenated and fed to $softmax$ for sentiment classification.
However, we cannot obtain better result with such an attention model. The accuracy of this attention model is slightly lower than the standard LSTM model (around 65\%), which means that the attention component has a negative impact on the model. 
A potential reason might be that the attention based LSTM has larger number of parameters, which cannot be easily optimized with the small number of corpus.

\section{Related Work}
We briefly review existing studies on target-dependent sentiment classification and neural network approaches for sentiment classification in this section. 

\subsection{Target-Dependent Sentiment Classification}

Target-dependent sentiment classification is typically regarded as a kind of text classification problem in literature.
Therefore, standard text classification approach such as feature-based Supported Vector Machine \cite{Pang2002,Jiang2011} can be naturally employed to build a sentiment classifier.
%For example, Jiang et al. \shortcite{Jiang2011} manually design target-independent features and target-dependent features with expert knowledge, syntactic parser and external resources.
Despite the effectiveness of feature engineering, it is labor intensive and unable to discover the discriminative or explanatory factors of data. 
To handle this problem, some recent studies \cite{Dong2014a,Vo2015} use neural network methods and encode each sentence in continuous and low-dimensional vector space without feature engineering. 
Dong et al. \shortcite{Dong2014a} transfer a dependency tree of a sentence into a target-specific recursive structure, and get higher level representation based on that structure.
Vo and Zhang \shortcite{Vo2015} use rich features including sentiment-specific word embedding and sentiment lexicons.
Different from previous studies, the LSTM models developed in this work are purely data-driven, and do not rely on dependency parsing results or external sentiment lexicons. 
%The approach is capable of capturing the relatedness of target word with its context words when composing continuous representation of sentence. 

\subsection{Neural Network for Sentiment Classification}
Neural network approaches have shown promising results on many sentence/document-level sentiment classification \cite{Socher2013a,Tang2015}. The power of neural model lies in its ability in learning continuous text representation from data without any feature engineering. 
For sentence/document level sentiment classification, previous studies mostly have two steps. They first learn continuous word vector embeddings from data \cite{Bengio2003,Mikolov2013a,Pennington2014}. Afterwards, semantic compositional approaches are used to compute the vector of a sentence/document from the vectors of its constituents based on the principle of compositionality \cite{Frege1892}. 
Representative compositional approaches to learn sentence representation include recursive neural networks \cite{Socher2013a,Irsoy2014nips}, convolutional neural network \cite{Kalchbrenner2014,Kim2014}, long short-term memory \cite{Li2015a} and tree-structured LSTM \cite{Tai2015,Zhu2015}.
There also exists some studies focusing on learning continuous representation of documents \cite{Le2014,Tang2015,Bhatia2015,Yang2016hierarchical}.

\section{Conclusion}
We develop target-specific long short term memory models for target-dependent sentiment classification.
The approach captures the connection between target word and its contexts when generating the representation of a sentence. 
We train the model in an end-to-end way on a benchmark dataset, and show that incorporating target information could boost the performance of a long short-term memory model.
The target-dependent LSTM model obtains state-of-the-art classification accuracy. 

\section*{Acknowledgements}
We greatly thank Yaming Sun for tremendously helpful discussions. 
%We also thank the anonymous reviewers for their valuable comments.
This work was supported by the National High Technology Development 863 Program of China (No. 2015AA015407), National Natural Science Foundation of China (No. 61632011 and No.61273321).
According to the meaning given to this role by Harbin Institute of Technology, the contact author of this paper is Bing Qin.

\bibliographystyle{acl}
\bibliography{bibtex}

\end{document}